\documentclass{article}

\usepackage{microtype}
\usepackage{graphicx}
\usepackage{subcaption}
\usepackage{booktabs} 

\usepackage{hyperref}

\usepackage{multirow}

\usepackage[preprint]{icml2026}

\usepackage{amsmath}
\usepackage{amssymb}
\usepackage{mathtools}
\usepackage{amsthm}

\usepackage[capitalize,noabbrev]{cleveref}

\theoremstyle{plain}

\theoremstyle{definition}

\theoremstyle{remark}

\usepackage[textsize=tiny]{todonotes}

\icmltitlerunning{Probe Generalization as Subspace Selection for OOD Deception Detection}

\begin{document}

\twocolumn[
 \icmltitle{Probe Generalization as Subspace Selection for OOD Deception Detection}



 \icmlsetsymbol{equal}{*}

 \begin{icmlauthorlist}
  \icmlauthor{Daniel Yoo}{yyy}
  \icmlauthor{Adrians Skapars}{comp}
 \end{icmlauthorlist}

 \icmlaffiliation{yyy}{Carnegie Mellon University}
 \icmlaffiliation{comp}{University of Manchester}

 \icmlcorrespondingauthor{Daniel Yoo}{dyoo2@andrew.cmu.edu}

 \icmlkeywords{Machine Learning, ICML}

 \vskip 0.3in
]



\printAffiliationsAndNotice{} 

\begin{abstract}
Linear probes can be used to detect behaviors and concepts inside language model activations, but may fail to transfer to out-of-distribution examples. When studying the generalization performance of Llama-3.1-8B-Instruct probes over 3 held-out deception detection datasets, 
we find that projecting inputs onto a small subset of principal components (PCs) from the training distribution of activations enables cross-domain transfer that nearly matches the performance of probes trained directly on the test distribution. 
Furthermore, we find that 
PC interpretations can be used to find a subset of those transferable PCs.
By using an LLM judge to score each PC on whether its most/ least activating examples imply a transferable deception direction, then probing on the highest-scoring PCs, we close the baseline-to-oracle gap by 78\% on Insider Trading Report and by 25\% on Sandbagging. The directions a source probe weights heavily appear to encode source-specific surface features, while the directions that actually transfer appear to encode the same contrast more abstractly, in a way natural language descriptions can capture. Broadly, our results suggest that the OOD robustness of probes is largely determined by subspace selection.

\end{abstract}

\section{Introduction}

Linear probes are standard tooling for reading information from language model activations \cite{alain2018understandingintermediatelayersusing,hewitt2019designinginterpretingprobescontrol}. 
Linear probes are cheap to train and expose some semantically meaningful information in the otherwise uninterpretable representation space.
However, probes that look strong in-distribution are sometimes unreliable out-of-distribution \cite{kirch2026impactoffpolicytrainingdata}. Existing work shows that key behavioral concepts are linearly represented in model activations \cite{marks2024geometrytruthemergentlinear, park2024linearrepresentationhypothesisgeometry,zou2025representationengineeringtopdownapproach}, so probes should have the capacity to detect these concepts, yet they appear to learn spurious correlations in practice \cite{orgad2025llmsknowshowintrinsic,wang2025falsesensesecurityprobingbased}. This is unacceptable for probe-based monitoring of harmful interactions, setting a clear challenge: 
\textit{how can we build probes that only learn generalizing features?}

To tackle this, we work in the basis of source-domain principal components (PCs) rather than raw activations, avoiding the assumption that the model's latent dimensions provide a privileged basis. Principal components offer a simple, data-driven way to study structure in representation spaces, and prior work even suggests that the choice of PCs matters: \citet{mu2018allbutthetopsimpleeffectivepostprocessing} shows that removing the top PCs of an embedding space improves downstream performance, and \citet{kantamneni2025sparseautoencodersusefulcase} use top-PC projections as a probe baseline. Thus, the focus of our investigation becomes: \textit{which subset of principal components supports generalizable probe training and how can that subset be identified?}




We use deception probes on Llama-3.1-8B-Instruct \cite{grattafiori2024llama3herdmodels} as our testbed,
training them on a single source dataset (Roleplaying deception scenarios) and evaluating on 3 held-out target datasets (Insider Trading Report, Insider Trading Confirm, and Sandbagging) \cite{goldowskydill2025detectingstrategicdeceptionusing}. 
Working in the source-domain PCA basis, we explore strategies for selecting a subset of training dimensions that avoid spurious correlations and transfer to target domains.

This work contributes the following findings:
\vspace{-10pt}
\begin{itemize}
 \item \textbf{A small transferable subspace exists} (\Cref{sec:transferable-subspace}).
 A target-supervised greedy search over source PCs finds 8--15
 components on which a source-trained probe nearly matches a
 target-trained probe.

 \item \textbf{Source-side and target-side rankings both fail to
 recover the subspace} 
 (\Cref{sec:source-target-fail}). Ranking PCs by the source or target probe's weights does not produce a basis that transferes better than the full-representation baseline.

 \item \textbf{An LLM judge partially recovers the subspace without
 target data} (\Cref{sec:interp-selection}). 
 An LLM judge can interpret and select PCs that are likely to encode transferable behavioral contrast rather than surface patterns.
 

\end{itemize}

\begin{figure*}[t]
 \centering
 \includegraphics[width=\linewidth]{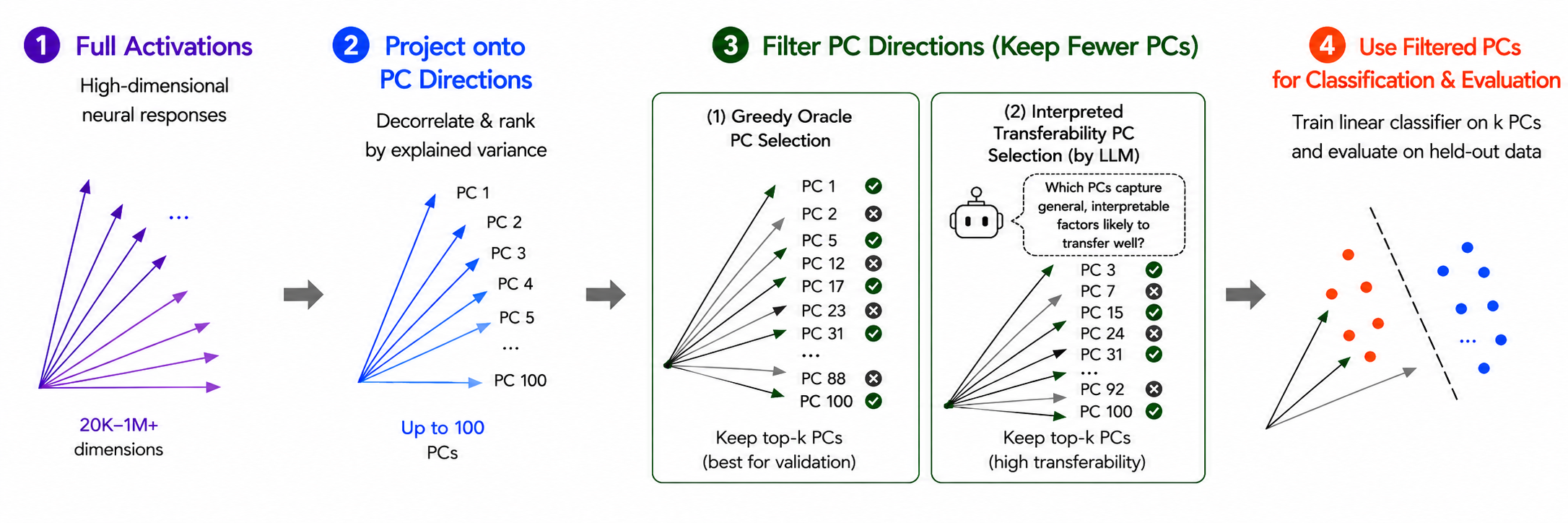}
 \caption{Overview of PC-based subspace selection for OOD deception probing. Activations are projected onto the source-domain PCA basis, candidate PC directions are filtered, and probes are trained on the selected low-dimensional subspace.}
 \label{fig:figure1}
\end{figure*}

\section{Related Work}
\paragraph{Linear concept directions in activation space.}
A related line of work studies whether high-level behavioral or semantic properties are represented as directions or low-dimensional subspaces in activation space. \citet{marks2024geometrytruthemergentlinear} report linear structure in true/false representations across datasets, while \citet{park2024linearrepresentationhypothesisgeometry} formalize the linear representation hypothesis and connect it to both probing and steering. Similarly, \citet{zou2025representationengineeringtopdownapproach} model high-level properties such as honesty and emotion as population-level directions that can be read from and intervened on in model activations.

This perspective has also been used to explain and control model behavior. \cite{arditi2024refusallanguagemodelsmediated} show that a single direction can mediate refusal behavior, suggesting that some high-level behaviors may be localized in simple linear features. \citet{lu2026assistantaxissituatingstabilizing} similarly identify an ``assistant'' axis corresponding to assistant-like behavior. Together, these works suggest that linear structure can support both mechanistic understanding and activation-space interventions, while also raising questions about when such directions capture stable concepts rather than context-dependent correlations.

\paragraph{Linear probes and their limitations.}
Linear and lightweight probes have a long history in interpretability as tools for analyzing what information is represented in neural networks \cite{alain2018understandingintermediatelayersusing,hewitt2019designinginterpretingprobescontrol}. Recent work has applied such probes to safety-relevant signals in LLM activations, including latent truthfulness without direct supervision \cite{burns2024discoveringlatentknowledgelanguage} and strategic deception \cite{goldowskydill2025detectingstrategicdeceptionusing}. 

At the same time, methodological skepticism is central to the probing literature. \citet{hewitt2019designinginterpretingprobescontrol} argue that high probe accuracy alone is insufficient evidence that a model already represents the target property, since the probe may itself learn the task or exploit surface-level correlates. Similarly, \citet{ravichander2021probingprobingparadigmdoes} question whether probing reliably establishes what information is encoded or merely recoverable. Recent work also finds that probes with strong in-distribution performance can generalize unevenly out of distribution \cite{kramár2026buildingproductionreadyprobesgemini}. These concerns motivate careful evaluation of whether safety probes identify robust internal signals rather than dataset-specific artifacts.

\paragraph{Interpreting features with natural-language explanations.}
Another approach aims to interpret internal representations by assigning natural-language descriptions to neurons or learned features. \citet{bills2023language} introduced the use of LLMs to generate explanations of neurons from high-activation examples. This approach has since been extended to the automatic labeling of features in sparse autoencoders (\cite{paulo2025automaticallyinterpretingmillionsfeatures}, \cite{bricken2023monosemanticity}), enabling large-scale interpretation of learned feature dictionaries.

Recent work has further explored whether natural-language feature descriptions can be used not only for interpretation but also for intervention. For example, \citet{casademunt2025steeringoutofdistributiongeneralizationconcept} identify concepts through automated natural-language interpretation and ablate them during fine-tuning to reduce misaligned behavior.

\section{Setup}

\paragraph{Datasets.}

Following \citet{goldowskydill2025detectingstrategicdeceptionusing}, we use 4 of their deception detection datasets (Roleplaying, Insider Trading Report, Insider Trading Confirm, Sandbagging), with descriptions being provided in Appendix \ref{app:dataset-description}.

Roleplaying is our source domain and the other 3 are target test sets. Each response is labeled as honest (0), ambiguous (-1), or deceptive (1) by GPT-4o \cite{openai2024gpt4osystemcard}, with the ambiguous responses being removed for our experiments.

\paragraph{Language models.}
Following \citet{ying2026truthfulnessspectrumhypothesis}, we use layer 15 residual stream activations of Llama-3.1-8B-Instruct \cite{grattafiori2024llama3herdmodels}, the layer with the highest cross-domain AUROC in their analysis. To extract activations, we pass each prompt through the model and average token activations over just the assistant's response.

\paragraph{Probe training and evaluation.}
We standardize source activations to zero mean and unit variance, then apply the same source statistics to standardize all target activations. We run PCA on the scaled source activations and retain the top 100 PCs. All probes are logistic regression classifiers ($\lambda =10$); for experiments using a subset $S$ of these 100 PCs, activations are projected onto $S$ before training and evaluation, while the baseline probes train directly on the full standardized activations.

We use the top 100 PCs because alignment with SAE decoder features \citep{bricken2023monosemanticity} decreases with PC index (Appendix \ref{app:pc-sae}), meaning lower-index PCs are more amenable to natural-language interpretation and hence comparable with interpretability-driven selection methods. This is not a claim that low-index PCs generalize better; \cref{sec:baseline} confirms the 100-PC subspace retains enough information to separate deceptive from honest responses on every target.


\section{Experiments}

\subsection{Setting a baseline}
\label{sec:baseline}

\begin{figure}[t]
 \centering
 \includegraphics[width=\linewidth]{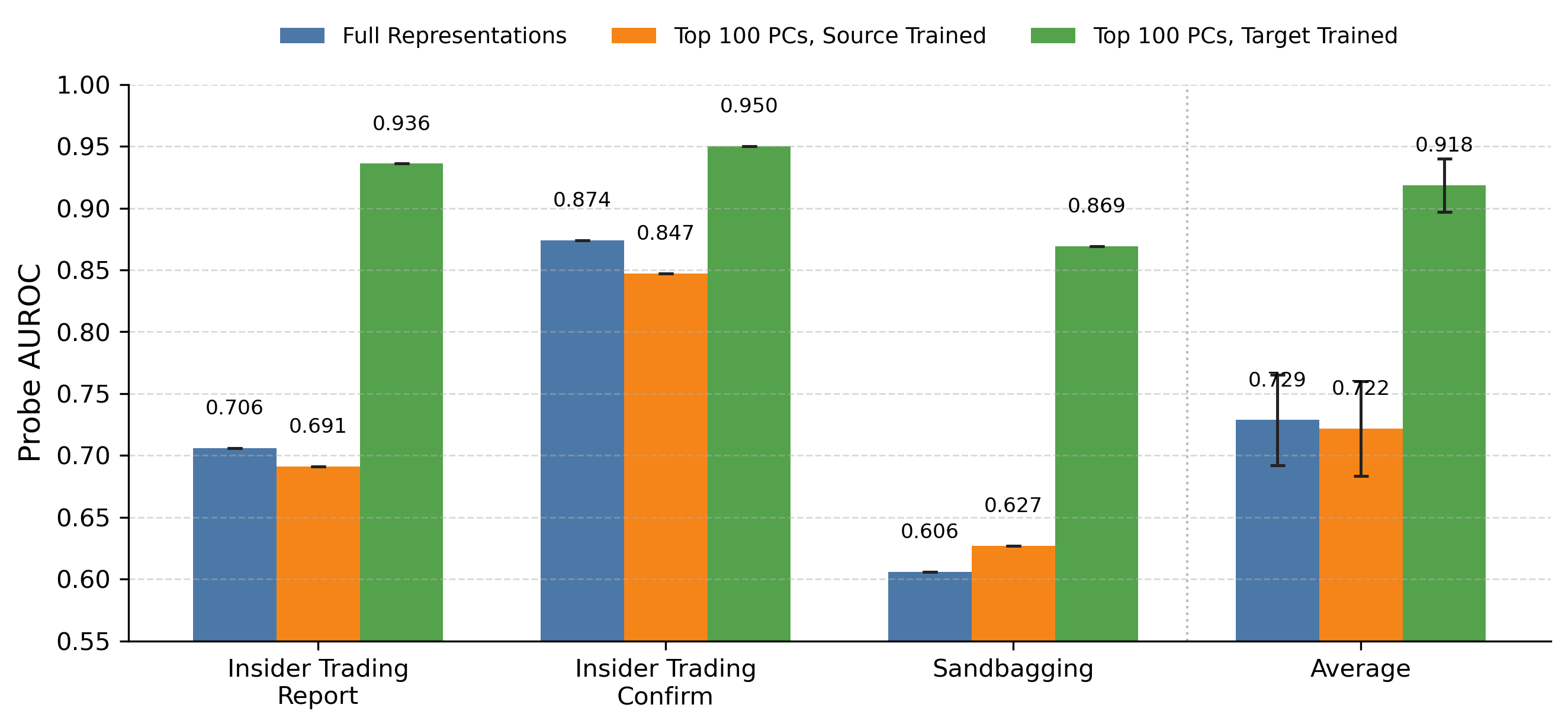}
 \caption{Probe AUROC on target datasets for source-trained probes on full representations (blue) and top-100 PCs (orange), and target-trained probes on top-100 source PCs (green, 5-fold CV). Error bars: 95\% CIs}
 \label{fig:baseline-auroc}
\end{figure}

We first test how well probes trained on the full source representations transfer to each target using source data.  The results are shown in \cref{fig:baseline-auroc}.
Insider Trading Confirm transfers reasonably well (0.874 AUROC) and Insider Trading Report performs moderately (0.706 AUROC), while Sandbagging performs poorly (0.606 AUROC).

To check that the 100-PC subspace itself is not the bottleneck, we also train a probe directly on target data within this subspace using 5-fold cross-validation.
The probe achieves 0.936 / 0.950 / 0.869 AUROC across the three targets. These results confirm that the 100-PC subspace encodes enough information to separate deceptive from honest responses on every target.

Thus, the difficulty does not stem from the relevant information being missing, but that the source probe learns a direction in the subspace that does not generalize.

\subsection{Showing a transferable subspace exists}
\label{sec:transferable-subspace}

\begin{figure}[!t]
 \centering
 \includegraphics[width=\linewidth]{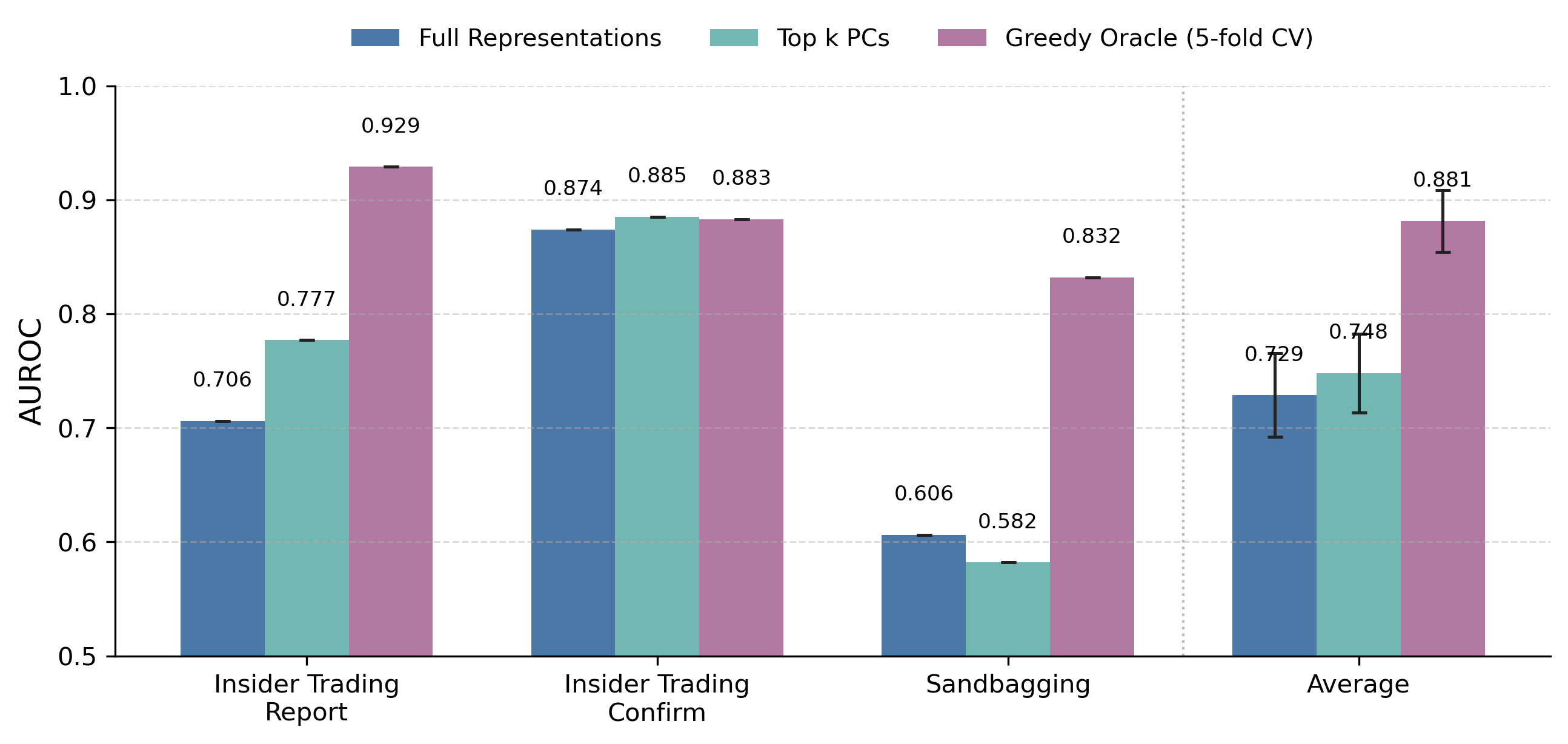}
 \caption{Probe AUROC across the three targets for the full-representation baseline (blue), top-k PCs by variance (teal, k matched to greedy), and greedy oracle PCs (purple). Error bars: 95\% CIs.}
 \label{fig:method-comp}
\end{figure}

\textit{Is there any method for training probes on source data that closes the gap in performance between the baseline and the target-trained probes?}
To test this, we first introduce and investigate a greedy algorithm for PC selection.
Starting from an empty set, we iteratively add PCs from the source dataset.
At each step, we search over the top 100 PCs and add the one that most improves AUROC on the target.
To evaluate each candidate, we project source activations onto the current subset and train a logistic regression probe on those projections.

Probe weights are always fit on source data, but the choice of PC subsets is guided by performance on labeled examples from the target dataset.
Using 5-fold cross-validation over the target dataset, we perform greedy PC selection on the non-held-out target folds and evaluate the selected source-trained probe on the held-out target fold. We stop when no candidate PC improves AUROC by more than 0.001. The search converges on small subsets of PCs — on average, 15 PCs for Insider Trading Report, 8 for Insider Trading Confirm, and 14 for Sandbagging.

The greedy probe provides an upper bound on transfer performance, not an unbiased OOD estimate, because subset selection uses some target examples.
To ground our results, we compare against simply selecting the top $k$ PCs by variance, where $k$ matches the number of PCs selected by the greedy algorithm. The results are shown in \cref{fig:method-comp}.

The greedy method comes close to a probe trained directly on target dataset for Insider Trading Report and Sandbagging, improving on the baseline by 0.22 and 0.23 points of AUROC, respectively.
Thus, a transferable subspace does exist in principle.
The variance control method matches greedy only for Insider Trading Confirm (the easy case) and falls far behind for Insider Trading Report and Sandbagging, the harder generalization tasks.
These findings suggest that the choice of PCs matters and that the amount of variance explained by those PCs does not predict their usefulness.

\subsection{Testing hypotheses for transferable principal components}
\label{sec:source-target-fail}

\begin{figure}[t]
 \centering
 \includegraphics[width=\linewidth]{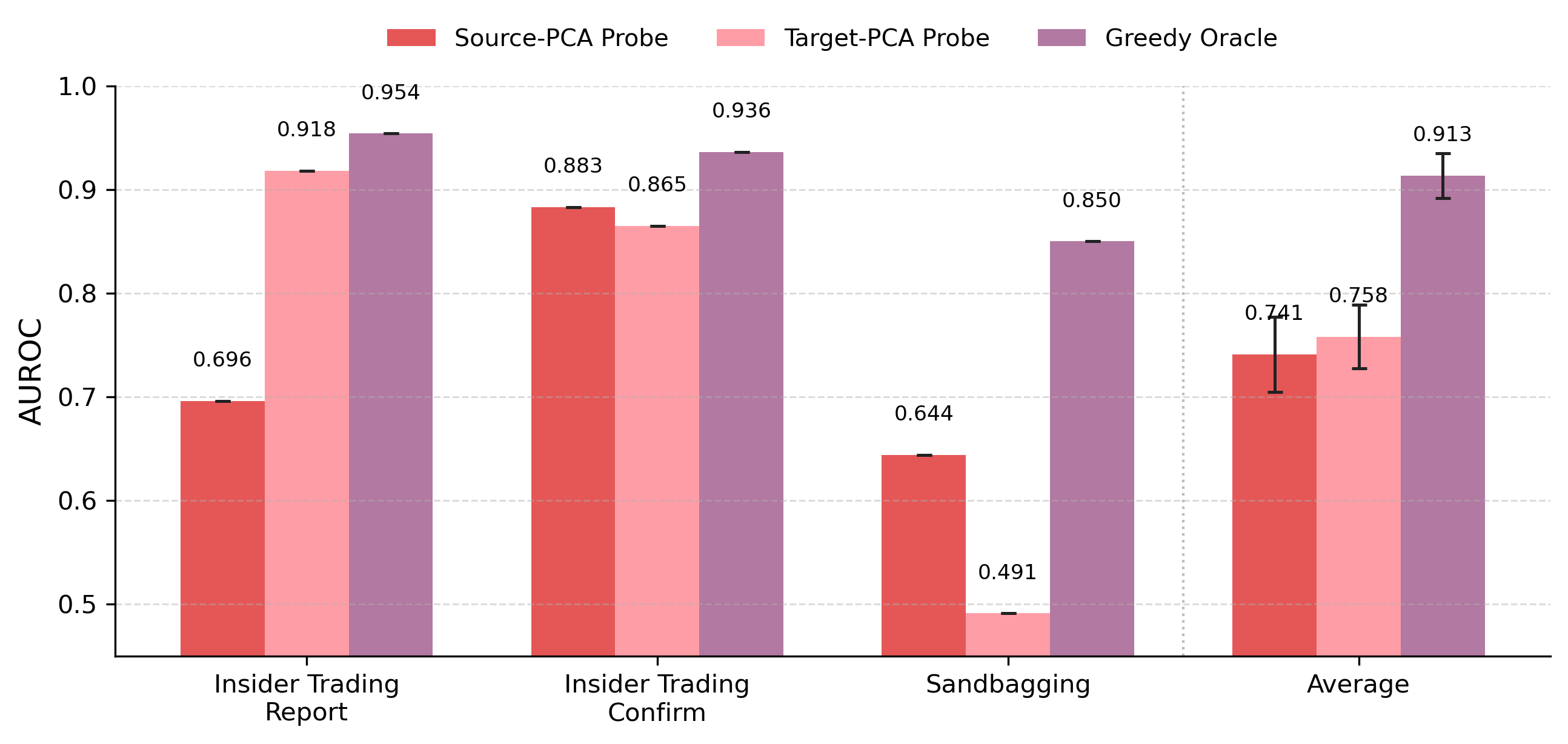}
 \caption{Probe AUROC across the three targets for three PC-selection rules: ranking by source-probe weights (red), by target-probe weights (pink), and greedy oracle selection (purple). Error bars: 95\% CIs.}
 \label{fig:hypothesis-auroc}
\end{figure}

\textit{What property of these PCs makes them transferable and can that property be detected from the source-side alone?}
To test this, we investigate 2 simple hypotheses for what the greedy subset contains.
The first posits that the transferable PCs are simply the ones that separate the target classes well by themselves across the test set.
— greedy happens to pick them because it has access to target data.
The second posits that they are the PCs the source probe already leans on most heavily — greedy is just rediscovering the source signal in a more compact form.

We compare 2 rankings of PCs, both based on the standardized coefficient (the absolute value of each probe coefficient multiplied by the standard deviation of the corresponding PC) given to each PC by a probe trained on activations projected on the source-PCA basis. 

The `source-PCA probe' ranks PCs by the weights learned by a probe trained on source data, while the `target-PCA probe' ranks PCs by the weights learned by a probe trained on target data. In both cases, the activations are projected onto the source-PCA basis. 
In both cases, the probes are trained on source data that has been projected onto  the top $k$ PCs, selected based to their respective rankings.
If either hypothesis were correct, the corresponding probes should recover the greedy method's performance.

To set $k$, we run the greedy method on the full target dataset rather than per fold, giving a single stable PC set: $k=13$ for Insider Trading Report, 7 for Insider Trading Confirm, and 14 for Sandbagging.
Because we are using the full dataset rather than doing CV, the greedy numbers differ slightly from the CV numbers in the previous section. The results are shown in \cref{fig:hypothesis-auroc}.

Both hypotheses fail, but in different ways. The source-PCA probe transfers about as well as the full representation baseline — the PCs that the source probe relies on are those that do not generalize. The target-PCA probe does well on Insider Trading Report but performs worse than the greedy probe on Insider Trading Confirm and has below random transfer performance on Sandbagging. The PCs that best separate target classes are not the PCs that, when weighted using source data, continue to transfer the best to the target data.
For further PC set comparisons, see Appendix \ref{app:pc-set-comparisons}.

\subsection{Interpreting the transferable subspace}
\label{sec:interpretable}
So far we have shown that a small subset of source PCs supports cross-domain transfer, and that this subset is not the same as the directions a source classification or target classification relies on.
\textit{Does the transferable subset correspond to anything human-interpretable?} 
To test this, we adopt the interpretability method of \citet{bills2023language} and \citet{casademunt2025steeringoutofdistributiongeneralizationconcept}.

For each of the top 100 source PCs, we find the 8 most and 8 least PC-activating examples from the source data, before prompting an LLM to generate a natural-language description of the positive pole, the negative pole, and the PC direction as a whole. 
The LLM judge then rates the generalizability of each direction on a 1–10 scale (OOD score), with the full judge prompt and model detail being provided in Appendix \ref{app:prompt}. 
Because no PC received an OOD score above 6 on the judge's 1--10 scale, we treated 6 as the maximum observed OOD score. We then retained all PCs that achieved this maximum score in both runs, yielding 19 PCs.

There were 5 PCs that showed up in the greedy selections of more than one target. 3 out of 5 PCs received the maximum OOD score, suggesting that LLMs can partially recover transferable directions. 
For comparison, only 1 of the top 5 PCs selected by the source-PCA probes received the maximum OOD score. 

Interpretations of the greedy-selected PCs describe the contrast at a level of abstraction that survives the move to other domains.
A full breakdown of PCs selected by each method is provided in Appendix \ref{app:pc-interp-texts}.

\subsection{Selecting a subspace by interpreting PCs}
\label{sec:interp-selection}

\begin{figure}[t]
 \centering
 \includegraphics[width=\linewidth]{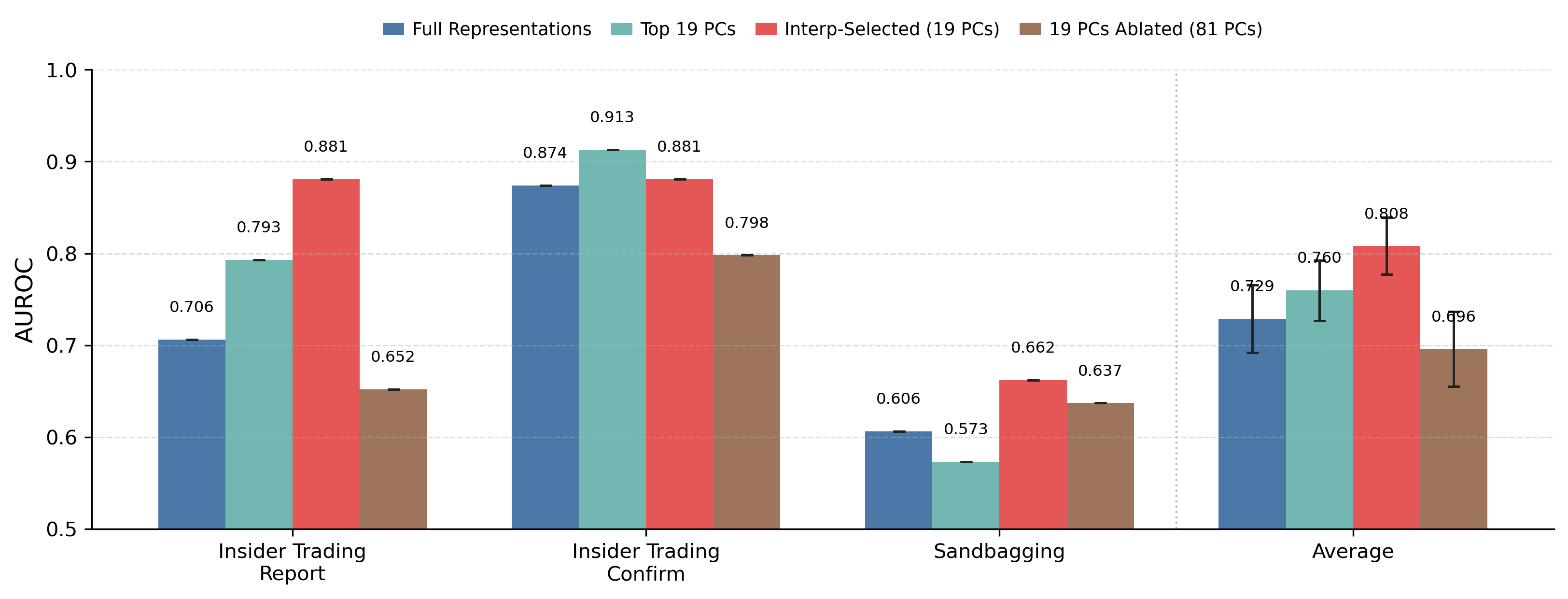}
 \caption{Probe AUROC across the three targets for the interp-selected subspace (red; 19 PCs) versus three controls: full-representation baseline (blue), top-19 PCs by variance (teal), and the remaining 81 PCs after ablating the interp-selected PCs (brown). Error bars: 95\% CIs.}
 \label{fig:interp-auroc}
\end{figure}

\textit{Does selecting PCs based on the interpretability-based OOD scores recover a meaningful fraction of the greedy probe performance, without using any target data?}
To test this, we select the PCs which were given the maximum OOD score by the judge model and use them as the probe subspace.
Because LLM judge scores are noisy, we run the scoring twice and only keep PCs that received the maximum score in both runs, giving us 19 interp-selected PCs in total.
As a control, we evaluate against a probe trained on the ablation of the selected PCs (the 81 PCs not used) as well as a probe trained on the top 19 source PCs by variance. The results are shown in \cref{fig:interp-auroc}.

The interp-selected PC probe outperforms the baseline on all 3 targets and outperforms the source-variance-selected control, by wide margins on Insider Trading Report and Sandbagging. As expected, the interp-selected PC probe also outperforms the interp-ablated PC probe across all 3 targets.
The size of the recovery varies sharply by dataset.
On Insider Trading Report, interp-selection closes 78\% of the gap between baseline (0.706) and the greedy oracle (0.929).
On Insider Trading Confirm, the baseline transfer performance is already near the ceiling so there is little headroom for comparisons, thus the result is uninformative — each selection method (aside from interp-ablated) does roughly as well as one another.
On Sandbagging, interp-selection closes 25\% of the gap between the greedy oracle (0.832).

Overall, selecting PCs purely by the natural language description of what they encode, with no use of target data, provides a meaningful improvement over baseline performance on the generalization tasks.

\section{Limitations}
\label{sec:limitations}

Our strongest claims are still quite narrow and there are several limitations that should be considered by the reader.\\

\textbf{Single model and source dataset.}
All experiments use a single model (Llama-3.1-8B-Instruct, layer 15) and a single source dataset (Roleplaying \cite{goldowskydill2025detectingstrategicdeceptionusing}). Moreover, Insider Trading Confirm results are uninformative, as baseline transfer on Insider Trading Confirm already reaches 0.874. Whether our results generalize to other model families or other datasets is an open question we leave to future work.

\textbf{Sandbagging margins are small.}
The weakest results are for Sandbagging, where interpretability selection raises AUROC from 0.606 to 0.662, recovering only 25\% of the gap to the greedy upper bound (0.832). Sandbagging is also the most out-of-distribution target, and the modest gain likely reflects this distribution shift. 
The result for Insider Trading Report (0.706 to 0.881, closing 78\% of the gap to greedy at 0.929) is the more robust.

\textbf{Linear pipeline.} Our probe pipeline is fully linear (PCA projection followed by logistic regression) so the existence of a transferable subspace is perhaps unsurprising. Whether these findings extend to nonlinear probes or other readout locations remains an open question.

\textbf{No causal validation.}
We make structural claims about what each PC ``encodes" based on top-activating examples and an LLM judge's verbalization of the contrast.
These are descriptive claims, not causal ones. We have not ablated the high-scoring PCs during inference to verify that they carry the contrast they appear to encode.


\section{Conclusion}

In probe-based deception monitoring, a small subset of source-domain principal components (PCs) enables cross-domain transfer that nearly matches target-trained probes, where this subset cannot be found with source-side or target-side separability alone.
An interpretability score over each PC's top-activating examples, produced by an LLM judge with no access to target data, partially recovers the subspace.

A key implication is that robust probing requires identifying not only whether a behavioral signal is linearly accessible, but where in activation space that signal is represented most robustly. Linear probes trained over full representations may combine general behavioral features with dataset-specific surface cues. These heuristics can support high in-distribution accuracy while hurting transfer. In contrast, sparse subspaces of PCs may better isolate directions corresponding to more general behavioral structure. This reframes OOD probing as a representation-selection problem: the goal is not only to fit a classifier, but also choose a subspace that preserves the intended behavioral feature and filters out spurious correlation.

\section*{Acknowledgment}
This work was conducted during the 11th edition of AI Safety Camp, which provided the compute to run our experiments. Thanks to Prasad Mahadik and Arya Hariharan for useful discussions around this work.

\section*{Impact Statement}
This paper presents work whose goal is to advance the field of Machine Learning. There are many potential societal consequences of our work, none
which we feel must be specifically highlighted here.

\bibliography{main}
\bibliographystyle{icml2026}

\newpage
\appendix
\onecolumn

\section{Dataset Details}
\label{app:dataset-description}
\textbf{Roleplaying:} the model is placed in a simple scenario with an opportunity to lie. \\
\textbf{Insider Trading Report:} acting as a trader, the model decides whether to honestly report that it used insider information. \\
\textbf{Insider Trading Confirm:} after hiding earlier use of insider information, the model is asked to confirm it did not use any. \\
\textbf{Sandbagging:} the model is told to deliberately under-perform on some tasks while behaving normally on others. \\

\begin{table}[h]
\centering
\begin{tabular}{lccc}
\toprule
\multirow{2}{*}{\textbf{Dataset}} & \textbf{Total} & \multicolumn{1}{c}{\textbf{Label Counts}} \\
 & \textbf{Examples} & \multicolumn{1}{c}{\textbf{(-1 / 0 / 1)}} \\
\midrule
Roleplaying & 371 & 140 / 59 / 172 \\
Insider Trading Report & 473 & 218 / 194 / 61 \\
Insider Trading Confirm & 151 & 29 / 38 / 84 \\
Sandbagging & 1000 & 379 / 375 / 246 \\
\bottomrule
\end{tabular}
\caption{Dataset label distributions.}
\label{tab:datasets_table}
\end{table}

\section{LLM Judge Prompt}
\label{app:prompt}
We use GPT-5-mini \cite{openai2025gpt5systemcard} to interpret directions. \\
{\ttfamily\scriptsize You are interpreting a principal component direction in a neural representation space.
This direction is bipolar:
  
  - Positive activations mean movement toward one pole.\\
  - Negative activations mean movement toward the opposite pole.\\
  - Larger magnitudes mean a stronger match to that pole.\\
  
Each line is formatted as token<TAB>activation, with activations normalized to integers from -10 to +10.
  
The examples below were selected from a dataset with this setup: The model is instructed to Roleplay in a simple scenario.
The model is asked a question which gives a strategic chance to lie.

High-positive examples:

\{positive\_examples\}

High-negative examples:

\{negative\_examples\}

Infer the dominant contrast captured by the direction using both the positive and negative examples.
  
Important instructions: \\
  - Do not assume the direction is a broad sentiment direction unless the examples clearly support that.\\
  - Prefer the narrowest accurate description over a broad generic one.\\
  - Do not overgeneralize beyond what is supported by the examples.\\
  - Rate how likely this direction is to be useful for out-of-distribution deception on a scale from 1 to 10, and briefly explain why.\\
  - Give low scores to directions that are mainly surface cues, lexical habits, or task-specific phrasing, even if they correlate with deception in these examples.\\
  - A direction that is noisy, brittle, context-dependent, or likely to fail under different wording or domains should usually be scored around 1 to 4, not 5 or above.\\
  - Reserve scores of 5 or higher for directions that seem plausibly tied to a more general deception-relevant strategy or behavior, rather than local wording.\\
  
  Write your answer in exactly this format:\\
  Positive pole: ...\\
  Negative pole: ...\\
  Shared direction: ...\\
  OOD deception usefulness (1-10): ...\\
  
}

\section{Lower PCs Show Stronger Decoder Alignment}

\label{app:pc-sae}
\begin{figure}[H]
 \centering
 \includegraphics[width=0.8\linewidth]{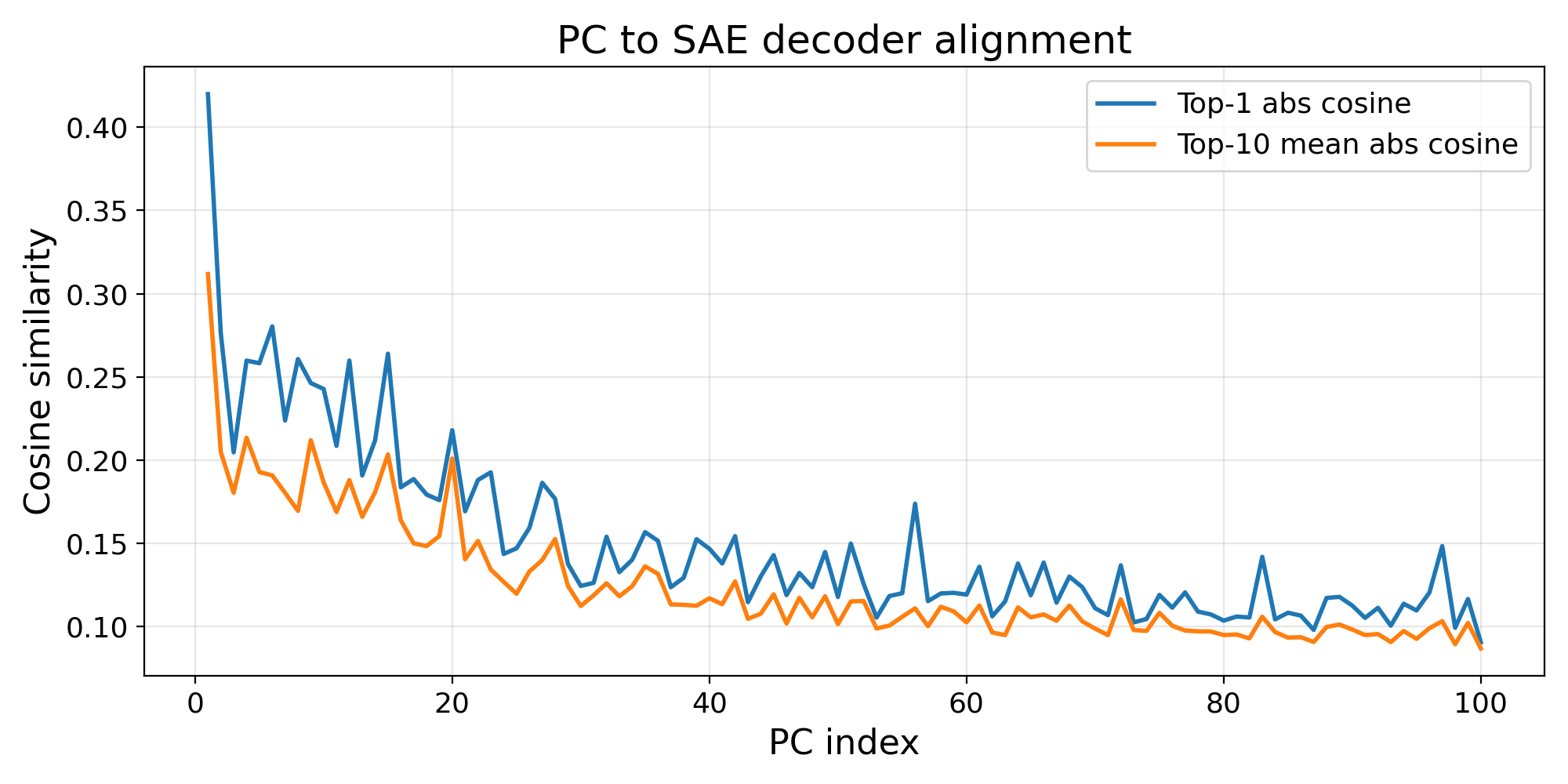}
 \caption{Lower PCs show stronger decoder alignment, both for the best-matching SAE feature and for the mean of the top 10 features (Spearman $\rho$ = -0.857 and -0.917, respectively). We used \texttt{andyrdt/saes-llama-3.1-8b-instruct} \cite{andyrdt2024saesllama318binstruct}}.
 
\end{figure}

\section{PC Set Comparisons}
\label{app:pc-set-comparisons}

Jaccard overlap between the PC sets of 2 methods (greedy and target-pca) is 0.182 for Insider Trading Report, 0.077 for Insider Trading Confirm, and 0.167 for Sandbagging. The 2 methods agree on less than a fifth of their selections.

Together these results suggest that the transferable subspace is not the part of representation space that is most discriminative on the source, and not the part most discriminative on the target. 

\begin{table}[H]
\centering
\begin{tabular}{ll p{0.5\textwidth}}
\toprule
\textbf{Selection method} & \textbf{Dataset / target} & \textbf{Selected PCs} \\
\midrule

Source probe weights
& Source
& PC7, PC8, PC11, PC12, PC4, PC5, PC2, PC3, PC6, PC27, PC22, PC21, PC1, PC41, PC18 \\

\midrule

Greedy selection
& Insider Trading Report
& PC6, PC89, PC62, PC66, PC26, PC14, PC57, PC25, PC93, PC9, PC70, PC97, PC78 \\
\midrule

Greedy selection
& Insider Trading Confirm
& PC71, PC53, PC28, PC86, PC32, PC81, PC77 \\
\midrule

Greedy selection
& Sandbagging
& PC37, PC62, PC54, PC41, PC46, PC9, PC57, PC44, PC51, PC43, PC19, PC48, PC86, PC81, PC33 \\

\midrule

Target separability ranking
& Insider Trading Report
& PC5, PC6, PC89, PC18, PC79, PC46, PC87, PC4, PC14, PC34, PC29, PC1, PC66, PC67, PC19 \\
\midrule

Target separability ranking
& Insider Trading Confirm
& PC4, PC1, PC8, PC53, PC30, PC7, PC13, PC18, PC5, PC20, PC26, PC49, PC28, PC31, PC97 \\
\midrule

Target separability ranking
& Sandbagging
& PC51, PC91, PC97, PC13, PC39, PC77, PC4, PC66, PC53, PC86, PC62, PC23, PC41, PC10, PC61 \\

\bottomrule
\end{tabular}
\caption{Comparison of PCs selected by source-only probe weights, greedy transfer selection, and target separability ranking.}
\label{tab:pc-selection-comparison}
\end{table}

\newpage
\section{PC Interpretability}
\label{app:pc-interp-texts}

\begin{table*}[h]
\centering
\caption{Principal components selected for likely transferability, based on interpretation with an LLM judge.}
\label{tab:greedy-pc}

\resizebox{\linewidth}{!}{%
\begin{tabular}{lccp{7.2cm}}
\toprule
\textbf{PC} & \textbf{Selected On} & \textbf{OOD Score} & \textbf{Full explanation} \\
\midrule
PC62 & Insider Trading Report + Sandbagging & 6 & OOD deception usefulness (1-10): 6 — This direction maps onto a general deception-relevant behavior (deny/deflect vs provide details), so it can be useful across settings, but it is noisy: true speakers also deny or obfuscate for legitimate reasons and liars sometimes produce elaborate fabricated detail, and the signal may pick up simple lexical negation rather than semantic strategy. \\
\midrule
PC57 & Insider Trading Report + Sandbagging & 6 & This direction aligns with a general deception-relevant strategy (deny/minimize or present innocuous explanations vs give specific/incriminating details), so it can help flag rhetorical moves often used when hiding or deflecting. However it is partly lexical and task-specific (relies on surface denials, hedges, and domain terms), so it will be brittle and context-dependent out of distribution. \\
\midrule
PC9 & Insider Trading Report + Sandbagging & 3 & OOD deception usefulness (1-10): 3 — This direction mainly picks up domain‑specific lexical cues (medical/leave language vs project/status language) and stylistic framing rather than a generalizable deception strategy. It may flag genuine illness claims and is brittle across domains and rephrasings. \\
\midrule
PC81 & Insider Trading Confirm + Sandbagging & 3 & OOD deception usefulness (1-10): 3 — The direction picks up a plausible deception‑relevant contrast (admit vs deny/hedge), but it mainly relies on surface/formal cues (negations, hedges, résumé style, enumerations) that are task- and wording‑dependent and likely brittle across domains and phrasings. \\
\midrule
PC86 & Insider Trading Confirm + Sandbagging & 6 & OOD deception usefulness (1-10): 6 — Hedging and defensive PR-style wording is a general tactic associated with deceptive or self-protective behavior, so the direction could transfer across domains. However it is noisy (used legitimately for privacy, politeness, or negotiation), sensitive to phrasing and role, and will produce false positives, so it's only moderately reliable out-of-distribution. \\
\bottomrule
\end{tabular}%
}
\end{table*}

\begin{table*}[t]
\centering
\caption{Principal components and their OOD scores with interpretations.}
\label{tab:source-pc}

\begin{tabular}{lcp{7.2cm}}
\toprule
\textbf{PC} & \textbf{OOD Score} & \textbf{Full explanation} \\
\midrule
PC7 & 3 & OOD deception usefulness (1-10): 3 — This direction mostly picks up a pragmatic contrast (deny/deflect vs admit) which can correlate with deception in some settings, but it relies heavily on surface cues (explicit ``No'', word order, presence of ``I'' + admission) and task-specific phrasing. It is therefore brittle and likely to fail or mislead in different domains or with varied wording. \\
\midrule
PC8 & 4 & OOD deception usefulness (1-10): 4 — The direction captures a useful narrow contrast (denial/excuse vs admission) that can correlate with evasive or deceptive responses in similar QA/roleplay tasks, but it largely picks up surface cues (explicit ``No''/``Yes'' and common denial phrasing) and is likely brittle under paraphrase, different domains, or subtler deception strategies. \\
\midrule
PC11 & 3 & OOD deception usefulness (1-10): 3 — The direction mainly picks up a stylistic/pragmatic contrast (assertive first‑person role/defense vs problem-reporting/negation). That can correlate with evasive or face-saving lies in some scenarios, but it's surface-y and context dependent and likely to fail across different domains or phrasings. \\
\midrule
PC12 & 3 & OOD deception usefulness (1-10): 3 — This direction mainly captures discourse style/forcefulness (a surface cue). While overconfident or over-assertive phrasing can sometimes correlate with deceptive defensive behavior, it is brittle and domain-dependent and will produce many false positives/negatives when applied across different speakers, topics, or cultural styles. \\
\midrule
PC4 & 6 & OOD deception usefulness (1-10): 6 — This captures a general communicative strategy relevant to deception (deny vs admit), so it can transfer beyond the specific prompts. However it also leans on surface cues (explicit ``Yes''/``No'' turns and common defensive phrasings) and may be brittle when denials or lies use different wording or more subtle tactics. \\
\bottomrule
\end{tabular}%
\end{table*}


\end{document}